\documentclass[lettersize,journal]{IEEEtran}
\usepackage{amsmath,amsfonts}
\usepackage{algorithmic}
\usepackage{pdflscape} 

\usepackage{algorithm}
\usepackage{array}
\usepackage[caption=false,font=normalsize,labelfont=sf,textfont=sf]{subfig}
\usepackage{textcomp}
\usepackage{stfloats}
\usepackage{url}

\usepackage{verbatim}
\usepackage{graphicx}
\usepackage{booktabs}
\usepackage{cite}
\usepackage{xcolor}
\usepackage{siunitx, xfrac}
\usepackage{glossaries}
\newcommand*\mygls[1]{%
  \protect\ifglsused{#1}{%
    \glsentryshort{#1}%
  }{%
    \glsentrylong{#1}%
  }%
}

\newacronym{XR}{XR}{Extended Reality}
\newacronym{HR2LLC}{HR2LLC}{High-Rate and High-Reliability Low-Latency Communications}
\newacronym[plural=APs,longplural=Access Points]{AP}{AP}{Access Point}
\newacronym{HMD}{HMD}{Head-Mounted Display}
\newacronym{QoE}{QoE}{Quality of Experience}
\newacronym{FoV}{FoV}{Field of View}
\newacronym{MAC}{MAC}{Medium Access Control}
\newacronym{LoS}{LoS}{Line-of-Sight}
\newacronym{SotA}{SotA}{State of the Art}
\newacronym{mmWave}{mmWave}{Millimeter-Wave}
\newacronym{IMU}{IMU}{Internal Measurement Unit}
\newacronym{ISAC}{ISAC}{Integrated Sensing and Communication}
\newacronym{THz}{THz}{Terahertz}
\usepackage{tabularx} 
\newacronym{Tx}{Tx}{Transmitter}
\newacronym{Rx}{Rx}{Receiver}
\newacronym{AoA}{AoA}{Angle of Arrival}
\newacronym{AoD}{AoD}{Angle of Departure}
\newacronym{BHI}{BHI}{Beacon Header Interval}
\newacronym{RIS}{RIS}{Reconfigurable Intelligent Surface}
\newacronym{6DoF}{6DoF}{Six Degrees of Freedom}
\newacronym{http}{HTTP}{hypertext transfer protocol}
\newacronym{COTS}{COTS}{commercial-off-the-shelf}
\newacronym{lldash}{LL-DASH}{Low-Latency Dynamic Adaptive Streaming Over HTTP}
\newacronym{dash}{DASH}{dynamic adaptive streaming over \mygls{http}}
\newacronym{webrtc}{WebRTC}{Web Real-Time Communication}
\newacronym{ietf}{IETF}{Internet Engineering Task Force}
\newacronym{moq}{MOQ}{Media over QUIC}
\newacronym{2d}{2D}{two-dimensional}
\newacronym{VMAF}{VMAF}{Video Multimethod Assessment Fusion}
\newacronym{MIMO}{MIMO}{Multiple-Input Multiple-Output}
\newacronym{MU-MIMO}{MU-MIMO}{Multi-User MIMO}
\newacronym{MCS}{MCS}{Modulation and Coding Scheme}
\newacronym{NR}{NR}{New Radio}
\newacronym{PPBP}{PPBP}{power per beam }
\newacronym{CSI}{CSI}{channel state information}
\newacronym{SNR}{SNR}{signal-to-noise ratio}
\newacronym{OFDM}{OFDM}{orthogonal frequency division multiplexing}
\newacronym{MPJPE}{MPJPE}{Mean Per Joint Position Error}
\begin{document}

\title{mmHSense: Multi-Modal and Distributed mmWave ISAC Datasets for Human Sensing}

\author{Nabeel Nisar Bhat, Maksim Karnaukh, Stein Vandenbroeke, Wouter Lemoine, Jakob Struye, Jesus Omar Lacruz, Siddhartha Kumar, Mohammad Hossein Moghaddam, Joerg Widmer, Rafael Berkvens, Jeroen Famaey


}



\maketitle

\begin{abstract}

This article presents \emph{mmHSense}, a set of open labeled mmWave datasets to support human sensing research within \gls{ISAC} systems. The datasets can be used to explore mmWave \gls{ISAC} for various end applications such as gesture recognition, person identification, pose estimation, and localization. Moreover, the datasets can be used to develop and advance signal processing and deep learning research on mmWave \gls{ISAC}. This article describes the testbed, experimental settings, and signal features for each dataset. Furthermore, the utility of the datasets is demonstrated through validation on a specific downstream task. In addition, we demonstrate the use of parameter-efficient fine-tuning to adapt ISAC models to different tasks, significantly reducing computational complexity while maintaining performance on prior tasks.
\end{abstract}

\begin{IEEEkeywords}
mmWave, integrated sensing and communicaiton, datasets, \gls{ISAC}, deep learning, testbeds, extended reality
\end{IEEEkeywords}

\section{Introduction}
Integrated Sensing and Communication (ISAC) \cite{liu2022integrated} enables communication networks to double as intelligent sensing systems, enabling advance human sensing applications. For instance, \gls{ISAC} can enable Wi-Fi routers to recognize human gestures in smart-home applications \cite{9606831}. In \gls{XR} setups, the same communication signals can track body pose without external cameras and hand-held controllers, creating more immersive and portable experiences. In autonomous vehicles, \gls{ISAC} can assist in precise localization \cite{9606831}. Since \gls{ISAC} reuses the existing wireless infrastructure, these sensing capabilities can be deployed widely without the high costs of dedicated sensors.

ISAC at higher frequencies such as millimeter waves (mmWave) \cite{gao2022integrated} provides added benefits through its high bandwidth and large antenna arrays, enabling fine range and angular resolution. This leads to more precise sensing, which can greatly enhance the performance of applications such as gesture recognition and localization. Although sub-6 GHz \gls{ISAC} has experienced rapid advancements in recent years, progress in mmWave \gls{ISAC} research has been comparatively slower. This slower progress is mainly due to the limited availability of \gls{COTS} mmWave devices, the absence of open-source tools for mmWave systems that give access to signal features such as \gls{CSI}, and the high cost associated with building custom experimental mmWave \gls{ISAC} platforms \cite{pegoraro2023disc}.

This work introduces \emph{mmHSense}\footnote{This work involved human subjects in its research. Approval
of all ethical and experimental procedures and protocols was granted by the
University of Antwerp Ethics Committee for the Social Sciences and Humanities (EA SHW)  under Application No. SHW\_2023\_313\_2. }, a collection of six mmWave \gls{ISAC} datasets designed to advance research on human sensing and bridge existing gaps in the field. The datasets incorporate various signal features, including mmWave \gls{CSI}, beam SNR, and \gls{PPBP}, collected using both COTS devices and custom experimental platforms (software-defined radio). Our datasets cover sub-6 GHz, mmWave Wi-Fi signals and 5G mmWave \gls{OFDM} signals, across bi-static and multi-static setups. Most existing mmWave datasets are radar-based. While DISC \cite{pegoraro2023disc} uses communication hardware, it is limited by its monostatic \gls{Tx}–\gls{Rx} configuration, which does not reflect real-world deployment scenarios, and it includes only five activities.

The key contributions of our work can be summarized as follows:
\begin{itemize}
    \item \textbf{Comprehensive mmWave \gls{ISAC} datasets:} We introduce \emph{mmHSense}, collection of mmWave \gls{ISAC} datasets to support a range of end applications, including gesture recognition, pose estimation, localization, and person identification. This work addresses the existing shortage of publicly available mmWave \gls{ISAC} datasets for human sensing.
    \item \textbf{Scalable COTS-based setups}: Most of our datasets are collected using \gls{COTS} Wi-Fi devices rather than dedicated mmWave radars. This makes the setups scalable and easily reproducible, ideal for real-world deployment, and also representative of fully integrated ISAC systems, where communication signals are leveraged for sensing tasks.
    \item \textbf{Signal features beyond \gls{CSI}}: Our datasets also include beam SNR and \gls{PPBP} as signal features, in addition to \gls{CSI}. Unlike \gls{CSI}, beam SNR and \gls{PPBP} are readily available as part of the standard sector sweep or beam sweeping process, requiring no additional overhead for extraction.
    \item \textbf{User, environment, and gesture diversity}: Unlike most existing mmWave \gls{ISAC} datasets, which are typically limited to a single user, a single environment, and a small set of gestures, our datasets encompass a wide variety of users, environments (ranging from isolated rooms to open corridors), and both predefined and natural gestures and poses. Furthermore, the data collection campaigns were conducted across multiple countries, including Spain, Belgium, and Sweden, ensuring greater geographic and environmental diversity
    \item \textbf{Distributed \gls{ISAC}:} Most \gls{ISAC} experiments focus on a single \gls{Tx}-\gls{Rx} setup, which at mmWave frequencies requires line-of-sight between users and devices. To overcome this limitation, one of our datasets, \emph{DISAC-mmVRPose}, explores distributed \gls{ISAC} with four \glspl{Rx} and a single \gls{Tx}, enabling more robust coverage and sensing capabilities.
    \item \textbf{Natural poses}: Moreover, most \gls{ISAC} datasets focus on predefined gestures and poses. In contrast, \emph{DISAC-mmVRPose} dataset captures realistic and natural gestures by allowing users to interact with a Virtual Reality (VR) game, resulting in more genuine poses and movements.
    \item \textbf{Multi-modal sensor fusion}: In most of our datasets, we incorporate multi-modal setups combining mmWave, sub-6 GHz Wi-Fi, and vision-based sensors. This enables pose estimation and supports advanced multi-sensor fusion techniques.
    \item \textbf{5G mmWave \gls{OFDM} signals:} While most existing mmWave \gls{ISAC} research relies on Wi-Fi signals, our \emph{5GmmGesture}  dataset explores \gls{OFDM} mmWave waveforms for gesture recognition. This serves as a primer for future 6G research, highlighting the potential of next-generation communication signals for ISAC.
    \item\textbf{Efficient fine-tuning with LoRA:} We explore the use of Low-Rank Adaptation to efficiently fine-tune ISAC models on new tasks without losing performance on the prior task. This significantly reduces training computational overhead enabling scalable and resource-efficient training for 6G foundation models.
    \item \textbf{Data and code:} Datasets are available via IEEE Dataport\footnote{https://ieee-dataport.org/documents/mmwavexr-multi-modal-and-distributed-mmwave-isac-datasets-human-sensing} and the code is available via GitHub\footnote{https://github.com/nisarnabeel/Multi-Modal-and-Distributed-mmWave-ISAC-Datasets-for-Human-Sensing/tree/main}.

\end{itemize}

\begin{table*}[t]
\centering
\caption{Overview of the mmHSense. 
GR = Gesture Recognition, PE = Pose Estimation, 
Loc = Localization, ID = Identification.}
\label{tab:datasets-overview-transposed}
\begin{tabularx}{\textwidth}{lXXXXXX}
\toprule
\textbf{Specs.} & \textbf{mmWGesture} & \textbf{mmWPose} & \textbf{mmW-GaitID} & \textbf{mmW-Loc} & \textbf{5GmmGesture} & \textbf{DISAC-mmVRPose} \\
\midrule
\textbf{Hardware} & Talon-AD7200 & wAP60Gx3 & wAP60Gx3 & wAP60Gx3 & Sivers EVK06002 & Sivers EVK06002 \\
\textbf{Signal feature} & Beam SNR & CSI amplitude & CSI amplitude & CSI amplitude & PPBP & CIR \\
\textbf{\#Hz mmWave} & 10 & 22 & 10 & 10 & 1540 & 2775 \\
\textbf{Task} & GR & PE & Person ID & Loc & GR & PE \\
\textbf{\#Subjects} & 3 & 3 & 20 & 20 & 8 & 8 \\
\textbf{Labels} & 10 classes & Regression & 8 classes & 20 classes & 8 classes & Regression \\
\textbf{Geometry} & 1 Tx–Rx & 1 Tx–Rx & 2 Tx–Rx pairs (X) & 2 Tx–Rx (X) pairs + Grid $4\times5$ & 1 Tx–1 Rx & 1 Tx–4 Rx \\
\textbf{Technology} & IEEE 802.11ad & IEEE 802.11ad & IEEE 802.11ad & IEEE 802.11ad & 5G NR & IEEE 802.11ay \\
\textbf{Aux Modality} & 5GHz CSI & Kinect skeletons & 5GHz CSI & 5GHz CSI & NA & Kinect skeletons \\
\textbf{Dataset size (samples)} & 854 & 2904 & 1318 & 2177 & 34496 & 32480 \\
\textbf{Duration (minutes)} & 220 & 110 & 40 & 60 & 72 & 14 \\
\bottomrule
\end{tabularx}
\end{table*}

\section{mmWave ISAC datasets: mmHSense }
\begin{figure*}[!t]
\centering
\includegraphics[width=\textwidth,trim = 0cm 1cm 0cm 0cm, clip]{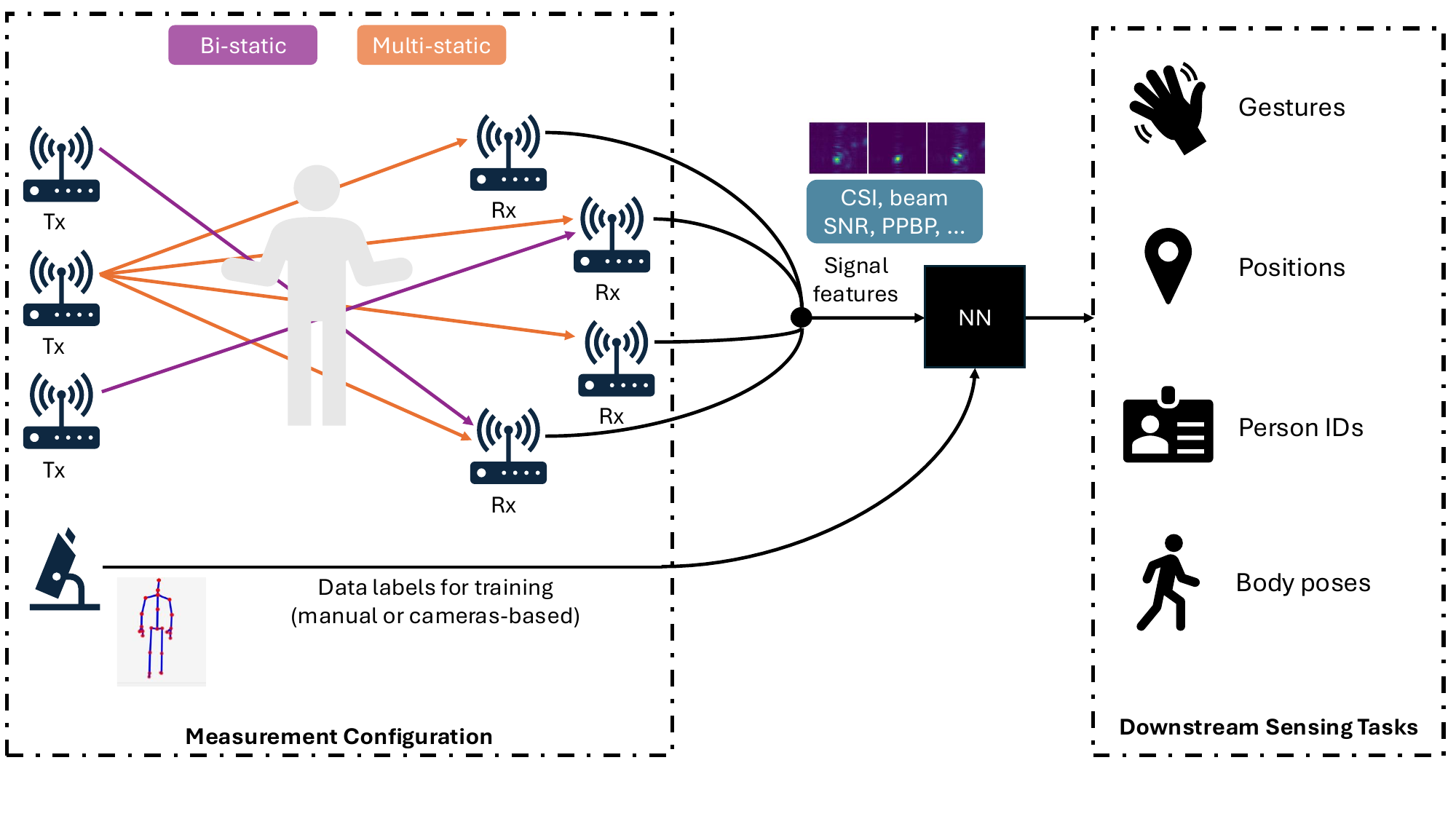}
\caption{Illustration of the mmWave ISAC-based experimental setup and processing pipeline. Multiple transmitter–receiver configurations (bi-static, two-pairs, and multi-static 1 \gls{Tx} - 4 \glspl{Rx}) are deployed to capture wireless signal features (e.g., \gls{CSI}, beam SNR, or \gls{PPBP}). The extracted features are processed by a neural network (NN) to enable diverse downstream tasks, including gesture recognition, localization, person identification, and pose estimation. Skeleton data and manual labels are incorporated only during the training phase.}
\label{fig:firsttwopapers}
\end{figure*}

\begin{figure}[!t]
\centering
\includegraphics[width=8.5cm]{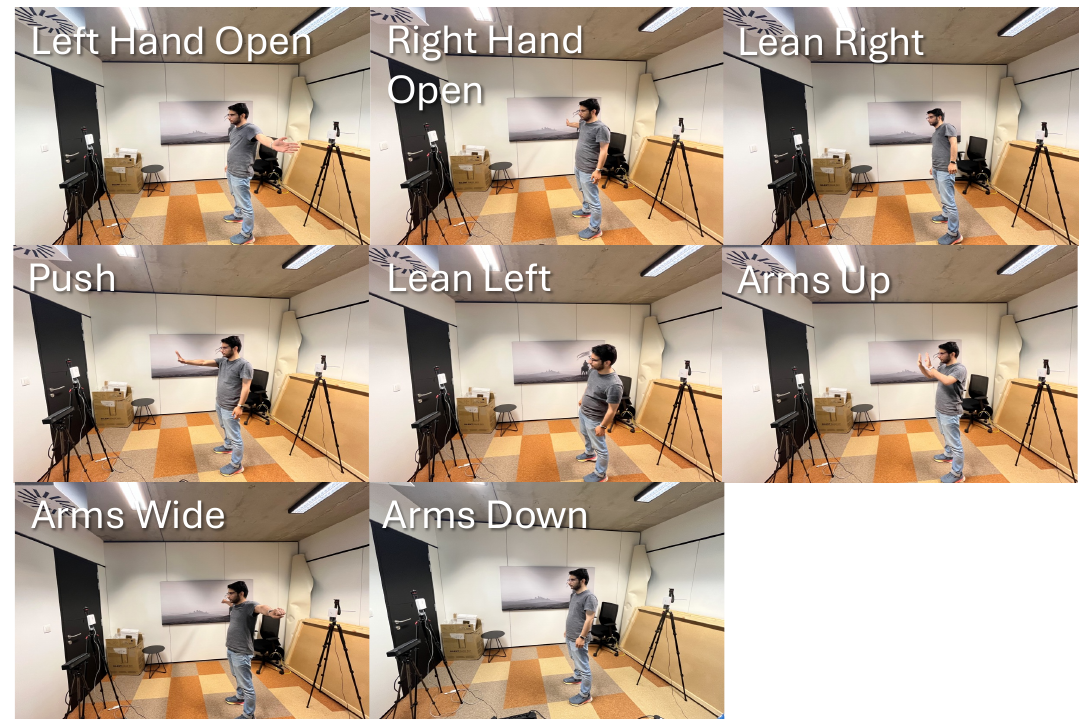}
\caption{Gestures and poses considered for Pose and 5GmmGesture dataset.}
\label{fig:gestures_list}
\end{figure}


Table \ref{tab:datasets-overview-transposed} shows the specifications (specs.) of different datasets
such as hardware used, signal feature, sampling rate (\#Hz) of signal feature, task, subjects, task type (labels), geometry, technology, auxiliary modality (Aux modality), dataset size and duration. Further, the experimental setup
illustrated in Figure \ref{fig:firsttwopapers} represents the general configuration underlying all datasets collected with mmWave wireless devices.
Bi-static and multi-static \gls{Tx}–\gls{Rx} configurations serve as the sensing backbone across different deployments. From these configurations, different signal
features, such as \gls{CSI}, beam SNR,
or \gls{PPBP}, are extracted depending on the
dataset. These features can be for example, directly fed to a neural network in an \emph{end-to-end} fashion, or undergo additional signal processing when required. Our datasets
include labels for several downstream tasks, including gesture
recognition, localization, person identification, and pose estimation.

\subsection{Gesture recognition} 
\paragraph{\textbf{mmWGesture}} In the first dataset, we use a pair of TP-Link Talon AD 7200 mmWave Wi-Fi devices that support the IEEE 802.11ad standard in a bi-static configuration (cf., Figure \ref{fig:firsttwopapers}). These devices periodically perform a sector sweep to determine the optimal beam for communication based on signal strength. Using an open-source tool\footnote{https://seemoo.de/talon-tools}, we get access to beam SNR for 36 such sectors. A user interrupts the line-of-sight between two devices functioning as \gls{Tx} and \gls{Rx} and performs a set of gestures and poses that trigger a change in beam SNR. 

We consider static gestures (poses) as well as dynamic gestures, including head rotations. We collect data in multiple environments, home and typical office space. In the home environment, we collected 125 minutes of activity data from a single user, resulting in 486 gesture examples. In the office environment, we recorded 221 gesture examples from three different users. Additionally, we conducted a rotation-based experiment in the home environment, where a single user performed 147 gestures while rotating 90 degrees, capturing orientation variability. Across all experiments, the total recorded activity time amounts to approximately 220 minutes. Note that the sampling frequency of the beam SNR is tied to the beacon interval of 102.4 ms, resulting in approximately 10 samples per second. More detailed on the performed gestures are provided in our previous work \cite{bhat2023gesture}.

The overall size of the dataset is $427000\times36$, where 36 represents the number of sectors and 427000 represents the number of samples. This can be for example reshaped into $854\times36\times500$, where 500 is the time index for each sample. Thus, a single gesture instance corresponds to a matrix capturing the temporal evolution of 36 features across 500 time steps.
\paragraph{\textbf{5GmmGesture}} This ISAC system consists of bi-static 5G mmWave nodes configured as a  \gls{Tx} and \gls{Rx}, implemented using two Sivers EVK06002 evaluation kits. These EVKs include front-end modules supporting frequencies in the 57–71 GHz range and are controlled via a programmable RF System-on-Chip (RFSoC).

The system transmits a 5G New Radio \gls{OFDM} waveform consisting of a Synchronization Signal Block (SSB) and randomized data over a single carrier comprising 792 subcarriers, spaced 960 kHz apart—resulting in a total bandwidth of approximately 760 MHz centered around 60 GHz. The transmitter continuously sends a 10 ms frame in compliance with 5G standards, composed of 10 subframes and 112 \gls{OFDM} symbols. Each \gls{OFDM} symbol includes a cyclic prefix (CP) of variable length and is processed using a 1024-point FFT. Beam sweeping is conducted over 50 transmit and 56 receive beams, with two \gls{OFDM} symbols transmitted per beam pair. The RFSoC is configured to sample the received signal such that exactly two \gls{OFDM} symbols—excluding their CPs—are captured for each \gls{Tx}-\gls{Rx} beam pair. The sampled received signal is then convolved with the corresponding transmit signal (excluding CPs) in the time domain. To ensure efficient on-chip processing, the convolution is performed using an FFT-based method, computed in blocks of up to 1024 samples. The resulting convolved signals are used to calculate the power per beam pair. This is done by computing the signal power and averaging across the samples and \gls{OFDM} symbols. Finally, the averaged \gls{PPBP} values are reshaped into a 50$\times$56 matrix to form the power grid, which serves as the input feature for gesture classification. This grid captures spatial energy distributions across the entire beam space, enabling fine-grained sensing and classification. 
We collect power per beam pair data from 8 diverse users performing 8 distinct gestures (cf., Figure \ref{fig:gestures_list}) for a total of 74.6 minutes of activity, with each gesture performed for 7 seconds and repeated 10 times. Unlike existing works that focus on \gls{CSI} or micro-Doppler, \gls{PPBP} is more robust (e.g., to
phase noise), and practical (e.g., not requiring advanced
synchronization). Moreover, the \gls{PPBP} measurements
can be easily extracted from the communication signals
themselves, rather than relying on dedicated sensing
packets. This approach is especially suitable
for \gls{ISAC}, entailing no overhead or additional resource
utilization.

The overall size of the dataset is $34496\times20\times50\times56$, where 34496 are the number of examples, 20 represents the time dimension, 50 and 56 represent the number of transmitter and receiver beams, respectively.
\subsection{Pose estimation}
\paragraph{\textbf{mmWPose}} This dataset is designed for skeletal body pose estimation. Data is collected using a single \gls{Tx}–\gls{Rx} pair, while a synchronized Kinect system provides the ground truth. The Kinect captures the 3D coordinates (x, y, z) of 25 human body joints. We employ MikroTik wAP 60Gx3 routers operating under
the IEEE 802.11ad standard, with CSI amplitude extracted
from 30 antenna elements using an open-source tool \footnote{https://github.com/IMDEANetworksWNG/Mikrotik-researchertools/tree/main}. The CSI is sampled at 22 Hz. Three participants perform eight distinct poses (cf., Figure \ref{fig:gestures_list}). Each pose was held
for a duration of 15 seconds, and the entire session involved
around 20 rounds of performing the set of eight poses, yielding a total of 110 minutes of activity. The multimodal pairing of mmWave and vision data facilitates cross-modal
learning and radio-to-vision translation.

The overall size of the dataset is $145200\times30$, where 145200 represents the number of samples and 30 represents number of antenna elements. This is reshaped in $2904\times30\times50$, where 50 represents the time index.
\paragraph{\textbf{D\gls{ISAC}-mmVRPose}}

\begin{figure}[!t]
\centering
\includegraphics[width=8cm]{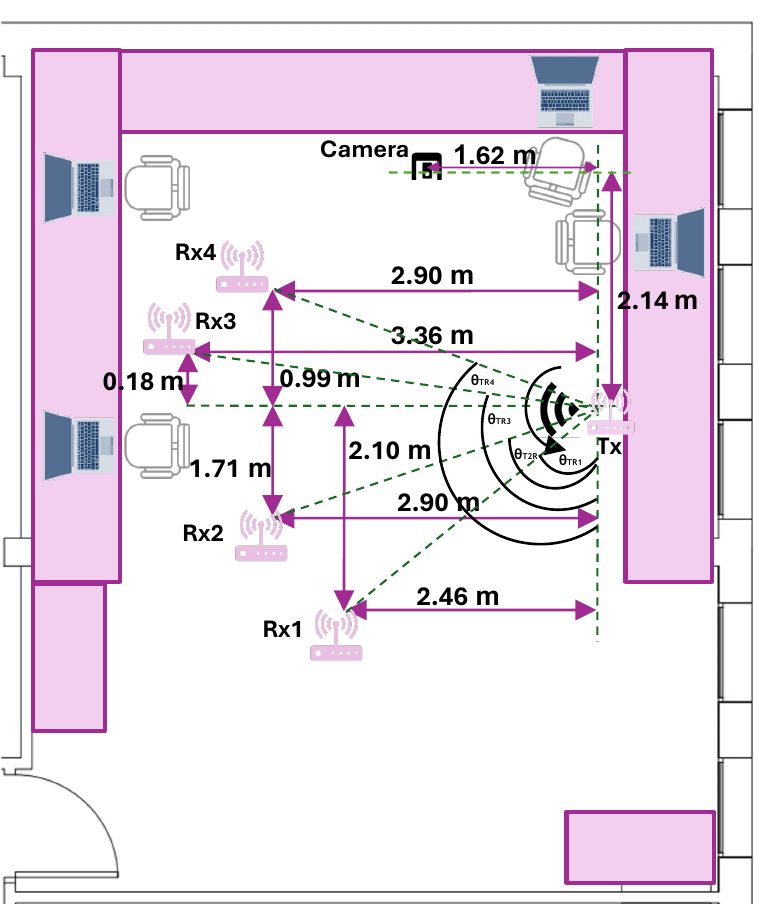}
\caption{Distributed ISAC setup VR Pose estimation: DISAC-mmVRPose dataset.}
\label{fig:imdea}
\end{figure}

We deploy a distributed \gls{ISAC} system based on the Mimorph testbed~\cite{MIMORPH}. For this data collection, the setup consists of a single mmWave \gls{Tx} and four mmWave \glspl{Rx}, all based on the Sivers Semiconductors EVK06002 phased-array antenna kits\footnote{Sivers Semiconductors, “Evaluation kit evk06002 for 57–71
ghz, unlicensed 5G mmwave (ieee 802.11ad),” https://www.
sivers-semiconductors.com/5g-millimeter-wave-mmwave-and-satcom/
wireless-products/evaluation-kits/evaluation-kit-evk06002/, 2025.} operating at a 60.48~GHz carrier frequency, and all connected to the same RFSoC board. The four \glspl{Rx} share a common 45~MHz local oscillator, ensuring coarse synchronization across all nodes. Each receiver is connected to the RFSoC through long cables, enabling their spatial distribution in the environment (cf., Figure \ref{fig:imdea}).

The \gls{Tx} sends training packets following the IEEE 802.11ay standard~\cite{Knightly_11ay}, each including training fields (TRN) transmitted with different beam patterns that uniformly sweep the azimuth range from $-45^{\circ}$ to $+45^{\circ}$. Each packet also includes a preamble composed of a short training field (STF) and channel estimation fields (CEF), which can be used for synchronization and impairment compensation. Packets are transmitted with an inter-frame spacing of 0.3~ms, allowing for Doppler extraction from human movement.

Ground-truth human poses are captured using a synchronized Microsoft Kinect, controlled from the same server that operates the Mimorph testbed. Data collection involves participants performing a series of gestures while interacting with a VR game, enabling the capture of realistic and naturalistic pose and gesture sequences. 8 volunteers participated to ensure diversity in the dataset. For each volunteer, we record approximately 105 seconds of VR interaction. 


Unlike most existing works, which are often limited to predefined gestures or a single \gls{Tx}–\gls{Rx} pair, our dataset captures unconstrained, natural movements from multiple \textit{synchronized} receivers, thereby providing multiple viewpoints of each action and enabling research into advanced multi-view data fusion techniques.

The dataset has dimensions \(32480 \times 4 \times 67 \times 1536\), where the first dimension represents the number of examples, the second dimension corresponds to the four \glspl{Rx}, the third dimension indexes time (\(67\) samples), and the last dimension encodes the beam–range grid (\(12 \times 128 = 1536\); 12 beams and 128 range bins).

\subsection{Person identification and localization fingerprinting }
\label{sec:mmw-multisense}
In these datasets, we employ MikroTik wAP~60Gx3 routers operating under the IEEE 802.11ad standard (same as mmWPose), with \gls{CSI} amplitude extracted from 30 antenna elements using an open-source tool\footnote{https://github.com/IMDEANetworksWNG/Mikrotik-researcher-tools/tree/main}. 

The mmWave CSI is sampled at 10 Hz. These datasets also have an additional auxiliary modality, 5GHz Wi-Fi, along with mmWave Wi-Fi. Since only CSI amplitudes are used, these datasets do not require additional preprocessing, such as \emph{phase sanitization}, making them particularly suitable for \emph{end-to-end deep learning approaches}.

\begin{figure}[!t]
\centering
\includegraphics[width=\columnwidth]{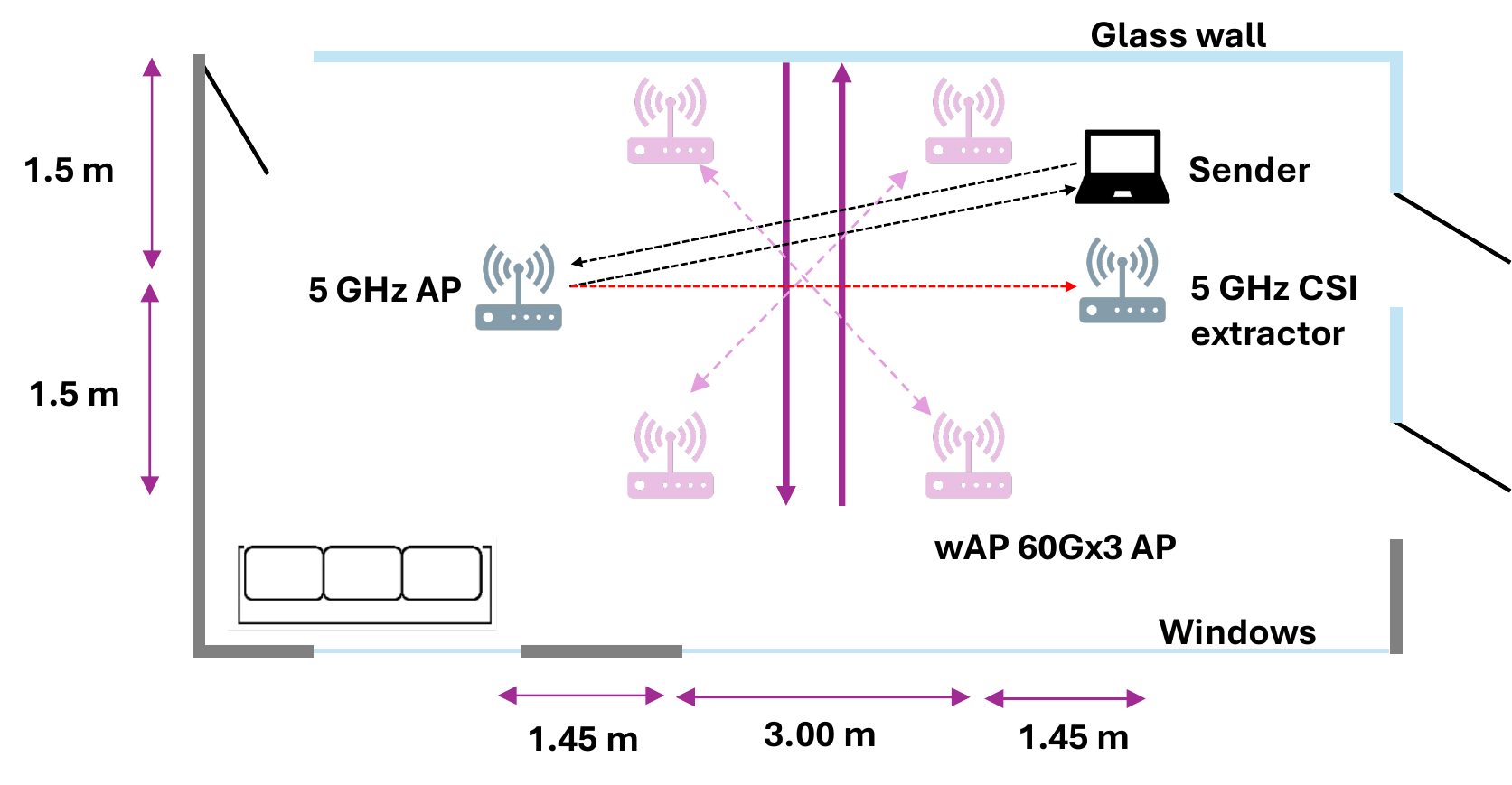}
\caption{Environment and experimental setup for mmW-GaitID and mmW-Loc.}
\label{fig:maxfloor}
\end{figure}

\paragraph{\textbf{mmW-GaitID}}
This dataset focuses on gait-based person identification. The mmWave setup consists of four wAP~60Gx3 units arranged as two independent Tx-Rx pairs in an X-shaped configuration. Figure \ref{fig:maxfloor} shows the experimental configuration of the setup. Arrows indicate the walking path. For cross-frequency comparison, a matching 5\,GHz system (IEEE 8012.11ac) is deployed using two ASUS~RT-AC86U routers (AP + passive monitor) with CSI extraction obtained via Nexmon\footnote{https://github.com/seemoo-lab/nexmon}. An Intel laptop sends ICMP echo requests to the AP, with the monitor capturing the CSI responses. The 5\,GHz CSI is downsampled to 10\,Hz to match the 60\,GHz data. 

Data collection spans three days with 20 participants (7/7/6), each walking for $\sim$2 minutes in a straight-line path with turns at each end. Background samples without any person present are also recorded. The 60\,GHz CSI has shape $\approx26{,}000\times30$ per device pair (30 representing number of antenna elements), while the 5\,GHz CSI has shape $\approx520{,}000\times52$, where 52 represents the number of subcarriers.

\begin{figure}[!t]
\centering
\includegraphics[width=8.5cm]{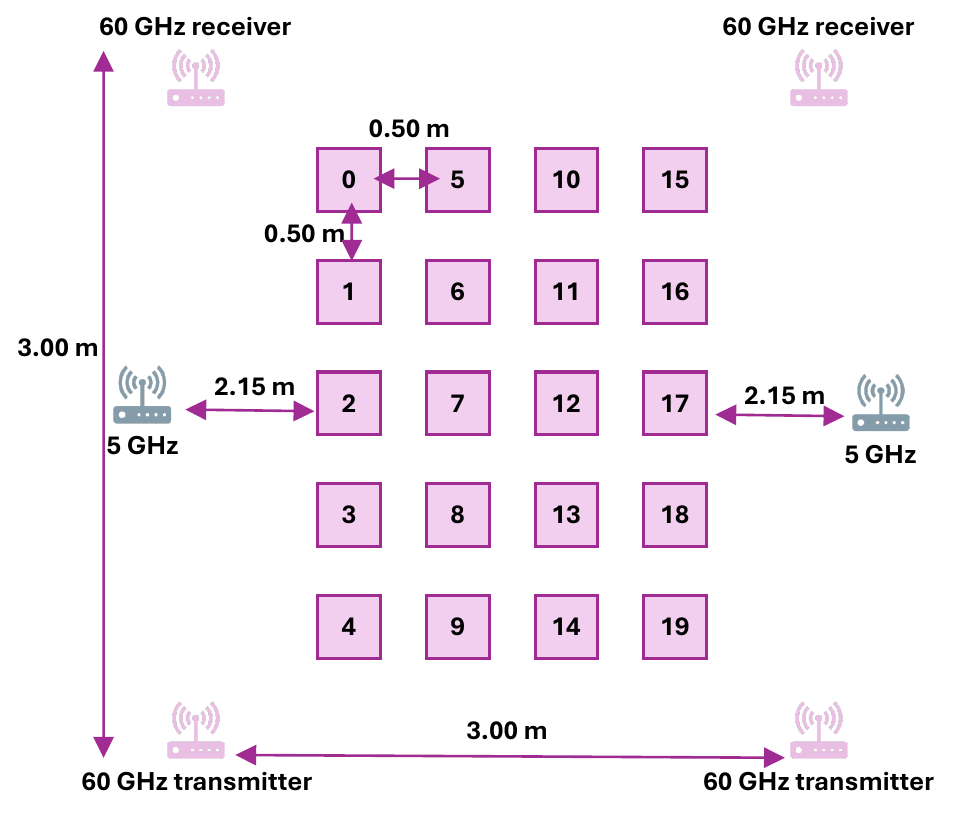}
\caption{Localization fingerprinting grid for mmW-Loc dataset.}
\label{fig:mmW-Loc}
\end{figure}
\paragraph{\textbf{mmW-Loc}}
This dataset addresses person localization fingerprinting over a discrete $4\times5$ spatial grid. The hardware configuration and the environment are identical to GaitID, with simultaneous recording of 60\,GHz and 5\,GHz CSI. Data is collected from 20 participants, each holding a fixed pose (standing still) for 10\,s at each of the 20 grid locations, shown in Figure \ref{fig:mmW-Loc}. This setup enables evaluation of cross-frequency localization performance and robustness across both spatial variation and participant diversity.

The overall size of the mmWave CSI is $21{,}770 \times 30$ per device pair. For the 5\,GHz CSI, the size is $1{,}404{,}060 \times 52$.

\section{Results}
\label{sec:results}
This section describes the validation of the datasets on specific end task. For the mmWGesture and the mmWPose datasets, the results are described in our previous works \cite{bhat2023gesture,bhat2023multi}.


\paragraph{5GmmGesture}
We split the data of PPBPs into train, validation and test sets. Here, the three sets contain PBPBs from all 8 users (in-domain settings). We leverage a ResNet-styled network \cite{he2016deep} to automatically extract features from PPBPs and map changes in PPBPs to the corresponding gestures, without any additional processing. We use ResNet because of skip connections and ability to reduce overfitting when training a deeper model. Table \ref{tab:results_all} shows the classification accuracy of the neural network for gesture recognition along with size, and GPU VRAM. Specifically, the model size represents the amount of RAM required to load the model parameters into memory. GPU VRAM indicates the approximate memory needed during training, which includes not only the model weights but also additional storage for gradients, optimizer states, and intermediate activations. These metrics provide a clear indication of the feasibility of performing such tasks on edge or resource-constrained devices, where memory and computational resources are limited. We can see that ResNet18 achieves 97.75\% accuracy on the test set while ResNet34 lags behind, due to complexity, resulting in slight overfitting.

However, when we evaluate the model in a cross-domain setting, we observe a notable drop in performance. In particular, the model is trained on data from seven users and tested on the held-out user. Since space is limited, we do not present the detailed results in a table or plot. The accuracy varies considerably, ranging from 54\% to 85\%, with an overall average of 71\%. This result underscores the challenges introduced by user-specific differences, commonly referred to as the \emph{domain gap}.





\paragraph{DISAC-mmVRPose} This dataset enables radio-to-vision translation, specifically mapping radio signals to 3D skeletons. It can also be leveraged for evaluating distributed sensing frameworks. This is a regression task, therefore, \emph{Mean Square Error} is used as loss function instead of \emph{Cross Entropy Loss} for classification tasks. Skeletons obtained from a Kinect camera serve as ground truth.

From the raw data obtained from four \glspl{Rx}, it is possible to obtain channel impulse response (CIR), and micro-doppler. In this work, our focus is on CIR. We show pose estimation performance evaluated on 8 individuals. Table \ref{tab:results_all} reports the \emph{\gls{MPJPE}}, which is a commonly used metric for pose estimation. \gls{MPJPE} measures the average Euclidean distance between the predicted and ground-truth 3D joint locations across all joints and frames.



We achieve an \gls{MPJPE} of approximately 6.2 cm using ResNet18 model, shown in Table 
\ref{tab:results_all}, which falls within the same magnitude as that achieved by camera-based computer vision methods. For example, Zhang et al. \cite{zhang2022mixste} in their recent work attain around 4.9 cm \gls{MPJPE} using spatio-temporal encoder, while in complex environments Gerats et al. \cite{gerats20233d} achieve \gls{MPJPE} in the 8–12.6 cm. Therefore, our mmWave-only system produces competitive accuracy, on par with camera-based computer vision systems, while relying entirely on mmWave CIR at test time.

 \begin{table*}[t]
\centering
\caption{Results across datasets: 5GmmGesture, DISAC-mmVRPose, mmW-Loc and mmW-GaitID.}
\label{tab:results_all}
\begin{tabular}{lcccc}
\toprule
\multicolumn{5}{c}{\textbf{5GmmGesture (Gesture Recognition)}} \\
Architecture & Accuracy (\%) & Trainable Params (M) & Size (MB) & GPU VRAM (MB) \\
\midrule
ResNet18 & 97.75 & 11.2 & 42.7 & 213.6 \\
ResNet34 & 96.15 & 21.3 & 81.2 & 487 \\
\addlinespace[0.5ex]

\multicolumn{5}{c}{\textbf{DISAC-mmVRPose (Pose Estimation)}} \\
Architecture & MPJPE (cm) & Trainable Params (M) & Size (MB) & GPU VRAM (MB) \\
\midrule
ResNet18 & 6.2 & 11.2 & 42.7 & 213.6 \\
ResNet34 & 6.0 & 21.3 & 81.2 & 487 \\
\addlinespace[0.5ex]

\multicolumn{5}{c}{\textbf{mmW-Loc (Task 1) and mmW-GaitID (Task 2)}} \\
Architecture & Accuracy (\%) & Trainable Params & Size (MB) & GPU VRAM (MB) \\
\midrule
\multicolumn{5}{c}{\emph{Standard Fine-Tuning (TCN Model)}} \\
TCN (Task 1) & 80.50 & 140K & 0.53 & 3.2 \\
TCN (Task 2) & 79.17 & 140K & 0.53 & 3.2\\
TCN (Task 1 after Task 2) & 9.15 & -- & -- & -- \\
\midrule

\multicolumn{5}{c}{\emph{Standard Fine-Tuning (Transformer Model)}} \\
Poolformer-s24 (Task 1) & 64 & 21M & 80.1 & 480 \\
Poolformer-s24 (Task 2) & 66.29 & 21M & 80.1 & 480 \\
Poolformer-s24 (Task 1 after Task 2) & 10.09 & -- & -- & -- \\
\midrule

\multicolumn{5}{c}{\emph{LoRA Fine-Tuning (Classification Head Only: TCN Model)}} \\
TCN (Task 1) & 81.88 & 140K & 0.53 & 3.2 \\
TCN (Task 2) & 73.80 & 2.2K  & 0.01 & 0.0050 \\
TCN (Task 1 after Task 2) & 81.88 & -- & -- & -- \\
\midrule

\multicolumn{5}{c}{\emph{LoRA Fine-Tuning (Adapters Across Network, Transformer Model)}} \\
Poolformer-s24 (Task 1) & 63.07 & 21M & 80.1 & 480 \\
Poolformer-s24 (Task 2) & 67 & 92K  & 0.35 & 2.1 \\
Poolformer-s24 (Task 1 after Task 2) & 63.07 & -- & -- & -- \\
\bottomrule
\end{tabular}
\end{table*}



\paragraph{mmW-Loc and mmW-GaitID}
In this experiment, we present results on mmWave CSI, where the GaitID dataset is reshaped to $1318\times60\times20$ and the mmW-Loc dataset to $2177\times60\times20$, with 60 representing the concatenation of 30 antenna elements from the two \glspl{Rx}, 20 represents the time index, while 1318 and 2177 represent the number of examples for the two datasets, respectively. Beyond reporting performance on identification and localization fingerprinting, we emphasize a critical challenge in ISAC-based sensing systems: catastrophic forgetting. Specifically, when a neural network trained on one task is subsequently retrained on another, its performance on the original task deteriorates significantly. This phenomenon poses a major obstacle for 6G foundation models, which are expected to generalize across diverse tasks while maintaining consistent performance on the previous tasks. 

\textbf{Classical Fine-tuning}:
Fine-tuning refers to training a pre-trained model on a new task to leverage the knowledge it has learned from the original task. Usually, some parameters or the entire model is retrained. However, this can lead to catastrophic forgetting, where the model loses knowledge of the original task while adapting to the new one.

To validate this effect, 
we first train a Temporal Convolutional Network (TCN) on an initial localization task using the mmW-Loc dataset. This pretrained model then serves as a baseline. We then fine-tune the same TCN model (all parameters) on a new person identification task using the GaitID dataset. Then, we re-evaluate performance on the Task 1 (inference) after training on Task 2 to measure the performance retention of the model. Table \ref{tab:results_all} shows the results of the model on the two tasks. While the model performs decently on the two tasks, however, fine-tuning the model makes it lose performance on the original task. We see that after fine-tuning, the performance of the model on the original task drops to 9.15\% for TCN and 10.09\% for transformer-based Poolformer-s24 model.

\textbf{LoRA-based fine-tuning}: To tackle this, we adopted Low Rank Adaptation (LoRA) \cite{hu2022lora}. LoRA is a parameter-efficient fine-tuning method that adapts pre-trained models by introducing small low-rank matrices to the frozen weights. By decomposing weight updates into two compact matrices, it greatly reduces the number of trainable parameters, enabling faster, more memory-efficient, and computationally cheaper fine-tuning compared to full model adaptation. 

We tackle the challenge of maintaining performance on Task 1 while adapting to Task 2 using two variants of LoRA-based fine-tuning. In the first approach, we freeze the entire pre-trained TCN and attach a lightweight low-rank adapter only to the classification head. During training, only the adapter parameters are updated, while the original network weights remain fixed. This strategy greatly reduces computational overhead and fully preserves Task 1 performance. However, since the majority of the network remains frozen and the adapter has limited capacity, the model struggles to fully adapt to Task 2, leading to a drop in accuracy on the new task (from 79.17\% to 73.80\%).

In the second approach, we adopt the PoolFormer transformer architecture for this setup, as the PEFT\footnote{https://huggingface.co/docs/peft/index} library used for LoRA fine-tuning requires standard architectures and cannot be directly applied to custom models. In this approach, the pre-trained PoolFormer backbone is kept frozen, and adapter layers are inserted into every convolutional block throughout the network. By distributing trainable parameters across multiple layers, the model gains greater representational capacity, enabling it to better capture features relevant to Task 2. This method improves Task 2 accuracy compared to adding adapters only to the classification head, while still preserving Task 1 performance and avoiding catastrophic forgetting. 

From Table \ref{tab:results_all}, we observe that LoRA significantly reduces both the number of trainable parameters and GPU memory requirements. For example, in the TCN model, the trainable parameters drop from 140K to only 2.2K, while the GPU VRAM usage decreases from 3.2 MB to just 0.005 MB, representing a 640× reduction. Similarly, for the Poolformer-s24, LoRA decreases trainable parameters from 21M to 92K and GPU VRAM from 480 MB to 2.1 MB, achieving a ~228× reduction. These drastic reductions are particularly important for training on mobile and edge devices, which have limited memory and computational resources. As models continue to scale toward 6G ISAC foundation models, potentially containing billions of parameters, full fine-tuning becomes computationally infeasible. In contrast, low-rank adapters make it possible to adapt large pre-trained models to new tasks efficiently, enabling real-time on-device training and rapid deployment in resource-constrained environments.

\section{Challenges and opportunities}
The main goal of this work is to show the validity of the datasets for mmWave ISAC-based human sensing. Future work may focus on developing new architectures, improving signal processing techniques, and exploring advanced learning methods, which remain open challenges for the research community. These datasets enable researchers to explore and address important challenges such as:
\paragraph{Domain adaptation} The results in Section \ref{sec:results} showed a significant reduction in accuracy on unseen user. Researchers can advance domain adaptation techniques using our datasets. Conventional domain adaptation methods, such as ADDA~\cite{tzeng2017adversarial} 
often fail in this scenario, as they rely on the assumptions that (i) the source and target domains share a common feature space and (ii) the conditional distribution of classes remains invariant across domains. 


However, for ISAC-based gesture recognition, this assumption often breaks down due to significant variations across users and environments, leading to a large domain gap. Therefore, such adaptation techniques may result in mismatches, for example source features of one gesture may be mapped to target features of another gesture. This data set provides a valuable benchmark for developing more robust domain adaptation techniques adapted to gesture recognition in ISAC systems.
\paragraph{Semi-supervised learning (SSL) and Foundation Learning} Collecting labeled data at mmWave frequencies is challenging due to limited research hardware and complex data collection processes. Moreover, existing unsupervised domain adaptation techniques often fail to transfer directly to ISAC systems. Our datasets provide a foundation for advancing SSL research in mmWave ISAC. Furthermore, these datasets open opportunities for unsupervised and self-supervised approaches, ultimately contributing to the development of mmWave foundation models \cite{du2024distributed} capable of learning from diverse and heterogeneous datasets.

\paragraph{Sensing and Communication Trade-off}
Our 5GmmGesture dataset could be used to analyze the trade-off between gesture recognition (sensing)
and communication performance, specifically examining how
the number of Tx and Rx beams, both in space and time,
impact the sensing performance and communication overhead
in ISAC.
\paragraph{Distributed ISAC}
Our dataset can also support research on split inference, where neural networks are distributed across multiple devices to reduce sensing data transfer bottlenecks and enable efficient, distributed multi-static ISAC \cite{zhuang2023integrated}.

\section{Conclusion}
In this paper, we introduced \emph{mmHSense}, a set of mmWave ISAC datasets designed to support a wide range of downstream tasks. These datasets capture diverse signal characteristics and include cross-modal settings. We described the testbed, experimental setup, and implementation details to ensure that our experiments are easily reproducible.

Furthermore, we demonstrated the utility of these datasets by evaluating their performance on several representative downstream tasks and highlighted key challenges encountered in real-world deployment. We hope this work serves as a valuable resource for the research community and invite researchers to leverage these datasets to advance deep learning techniques and address the open challenges in mmWave ISAC.
\section*{Acknowledgments}
This research was partially funded by the Research Foundation - Flanders (FWO) project WaveVR (Grant number G034322N) and European Commission through the Horizon Europe JU SNS project Hexa-X-II (Grant Agreement no. 101095759). Part of this work was supported by the European Union’s Horizon Europe programme
under the SNS-JU through Grant 101192521 (MultiX); the Comunidad de
Madrid through projects DISCO6G-CM (TEC-2024/COM-360) and TUCAN6-CM
(TEC-2024/COM-460) under ORDEN 5696/2024; project PID2022-136769NB-I00
(ELSA) funded by MCIN/AEI /10.13039/501100011033 / FEDER, EU. Nabeel Nisar Bhat is supported by an FWO SB PhD fellowship (Grant number 1SH5X24N).



\bibliographystyle{IEEEtran}
\bibliography{IEEEabrv,bibliography}

\begin{IEEEbiographynophoto}
{\bfseries Nabeel Nisar Bhat} is a Ph.D. student at the University of Antwerp and imec, Belgium. His research focuses on mmWave-based ISAC for human sensing applications.
\end{IEEEbiographynophoto}

\begin{IEEEbiographynophoto}
{\bfseries Maksim Karnaukh} obtained Masters Degree at Department of Computer Science, University of Antwerp.
\end{IEEEbiographynophoto}

\begin{IEEEbiographynophoto}
{\bfseries Stein Vandenbroeke} is a Master student at Department of Computer Science, University of Antwerp.
\end{IEEEbiographynophoto}
\begin{IEEEbiographynophoto}
{\bfseries Wouter Lemoine} is a PhD student at Department of Computer Science, University of Antwerp.
\end{IEEEbiographynophoto}

\begin{IEEEbiographynophoto}
{\bfseries Jakob Struye} is a postdoctoral researcher at University of Antwerp.
\end{IEEEbiographynophoto}

\begin{IEEEbiographynophoto}
{\bfseries Jesus Omar Lacruz} is a Senior Research engineer at Imdea Networks Institute, Spain.
\end{IEEEbiographynophoto}

\begin{IEEEbiographynophoto}
{\bfseries Siddhartha Kumar} is a senior engineer at Qamcom Research, Sweden.
\end{IEEEbiographynophoto}

\begin{IEEEbiographynophoto}
{\bfseries Mohammad Hossein Moghaddam} is a senior engineer at Qamcom research, Sweden.
\end{IEEEbiographynophoto}

\begin{IEEEbiographynophoto}
{\bfseries Joerg Widmer} is a Research Professor and Research Director at IMDEA Networks Institute, Madrid, Spain.
\end{IEEEbiographynophoto}
\begin{IEEEbiographynophoto}
{\bfseries Rafael Berkvens} is a Professor with the Department of Electronics-ICT at the University of Antwerp and a principal investigator with imec, Belgium.
\end{IEEEbiographynophoto}
\begin{IEEEbiographynophoto}
{\bfseries Jeroen Famaey} is a Professor at the University of Antwerp and imec, Belgium.
\end{IEEEbiographynophoto}

\vfill

\end{document}

I can definitely help with that. You've correctly observed that the number of trainable parameters for the LoRA model is a tiny fraction of the total parameters. This is the core benefit of using LoRA.

Based on the output you provided, here's the breakdown of the size:

Total model size (including frozen and trainable parameters): 80.03 MB

Total trainable parameters (LoRA adapters only): 92,464

To get the size of just the trainable parameters in megabytes, we can do a simple calculation. Since each parameter in PyTorch is typically a 32-bit floating-point number (4 bytes), we can calculate the size like this:

(92,464 parameters×4 bytes/parameter)/(1024×1024 bytes/MB)≈0.35 MB
As you can see, the size of the trainable LoRA parameters is only about 0.35 MB. This is a small amount compared to the full model size of 80.03 MB, which is less than 0.5

The main idea behind LoRA is to save and load only these small, trainable parameter matrices, which makes fine-tuning large models incredibly efficient.

\begin{IEEEbiography}
{\bfseries Nabeel Nisar Bhat} is a Ph.D. student at the University of Antwerp and imec, Belgium. His research focuses on mmWave-based integrated sensing and communication (ISAC) for human sensing applications.
\end{IEEEbiography}

\begin{IEEEbiography}
{\bfseries Maksim Karnaukh} is a [position, e.g., Ph.D. student or researcher] at [affiliation, e.g., University of Antwerp, imec, or another institute]. His research interests include [research areas].
\end{IEEEbiography}